\documentclass[10pt,twocolumn,letterpaper]{article}

\usepackage{wacv}
\usepackage{times}
\usepackage{epsfig}
\usepackage{graphicx}
\usepackage{amsmath}
\usepackage{amssymb}
\usepackage{xspace}

\usepackage{booktabs}
\usepackage{adjustbox}
\usepackage{paralist}
\usepackage{algorithm}
\usepackage[noend]{algpseudocode}
\usepackage{bm}
\usepackage{xcolor}
\usepackage{enumitem}
\setitemize{noitemsep,topsep=0pt,parsep=0pt,partopsep=0pt}

\usepackage[normalem]{ulem} 

\newcommand{\ours}{DualAdapt\xspace}

\DeclareMathOperator*{\argmax}{arg\,max}

\def\wacvPaperID{507} 

\wacvfinalcopy 

\ifwacvfinal
\def\assignedStartPage{9876} 
\fi

\ifwacvfinal
\usepackage[breaklinks=true,letterpaper=true,linkcolor=blue,citecolor=blue,bookmarks=false]{hyperref}
\else
\usepackage[pagebackref=true,breaklinks=true,letterpaper=true,colorlinks,linkcolor=blue,citecolor=blue,bookmarks=false]{hyperref}
\fi

\ifwacvfinal
\else
\fi

\begin{document}

\title{Federated Multi-Target Domain Adaptation}

\author{
Chun-Han Yao$^{1,2}$ \hspace{2mm}
Boqing Gong$^1$ \hspace{2mm}
Yin Cui$^1$ \hspace{2mm}
Hang Qi$^1$ \hspace{2mm}
Yukun Zhu$^1$ \hspace{2mm}
Ming-Hsuan Yang$^{1,2}$
\\
\\
$^1$Google Research \hspace{3mm}
$^2$UC Merced
}

\maketitle
\thispagestyle{empty}

\begin{abstract}
Federated learning methods enable us to train machine learning models on distributed user data while preserving its privacy.
However, it is not always feasible to obtain high-quality supervisory signals from users, especially for vision tasks.
Unlike typical federated settings with labeled client data, we consider a more practical scenario where the distributed client data is unlabeled, and a centralized labeled dataset is available on the server.
We further take the server-client and inter-client domain shifts into account and pose a domain adaptation problem with one source (centralized server data) and multiple targets (distributed client data).
Within this new Federated Multi-Target Domain Adaptation (FMTDA) task,
we analyze the model performance of exiting domain adaptation methods 
and propose an effective \ours method to address the new challenges.
Extensive experimental results on image classification and semantic segmentation tasks demonstrate that our method achieves high accuracy, incurs minimal communication cost, and requires low computational resources on client devices.
\end{abstract}

\vspace{-5mm}
\section{Introduction}
\vspace{-1mm}
\begin{figure}[t]
    \centering
    \includegraphics[width=.99\linewidth]{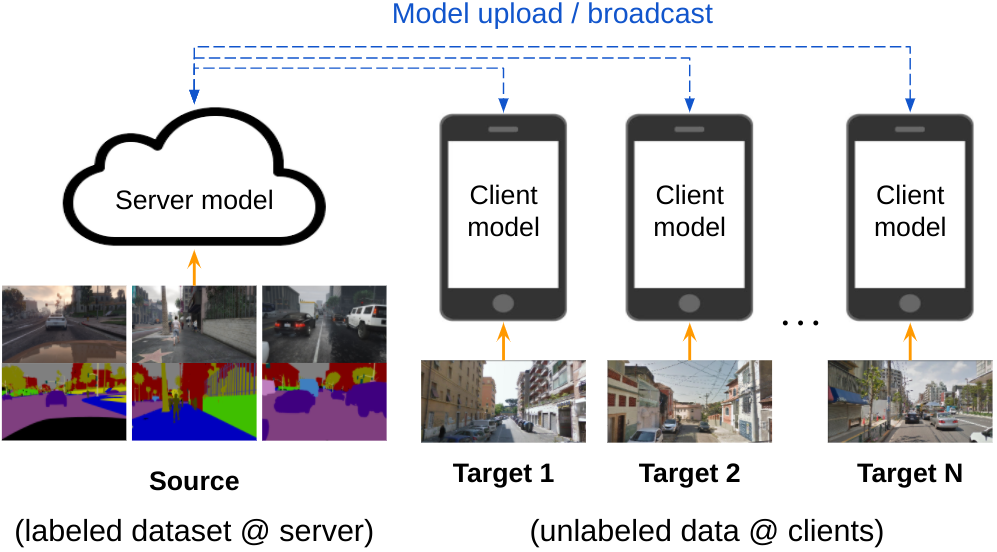}
    \vspace{1mm}
    \caption{\textbf{Federated Multi-Target Domain Adaptation (FMTDA), a new problem setting this work studies.}
    The client models are trained in a distributed, privacy-preserving manner and aggregated on the server.
    Unlike typical FL scenarios, we aim to train a model using unlabeled client data and a labeled dataset available only on the server.
    The server and clients are data providers with distinct characteristics, resulting in both server-client and inter-client domain gaps.
    Therefore, it naturally forms a problem of single-source-multi-target domain adaptation.}
\label{fig:intro}
\vspace{-2mm}
\end{figure}
%
%
%
%
Federated Learning (FL)~\cite{bonawitz2017practical, mohassel2017secureml, smith2017federated, mohassel2018aby3} aims to train a model using the data and computational resources on local devices to preserve privacy.
While most existing FL methods assume labeled client data at our disposal, the assumption may be impractical for numerous computer vision tasks as it is difficult to obtain the ground-truth annotations.
For instance, typical mobile phone users are unlikely to label object segmentation masks or bounding boxes on their abundant photos.
Although many large-scale datasets for such tasks have been created and labeled by the vision community, storing and using them on local devices can cause memory, communication, or computational overload. 
Furthermore, there usually exists a large domain gap between the curated dataset and on-device user data.

To deal with the scenarios discussed above, we study a practical FL setting in this work. 
The server hosts a large-scale labeled dataset, viewed as the source domain, and each client is considered a distinct target domain with unlabeled data.
The client data has different underlying distributions due to various factors, \eg, user habits, locations, and devices. 
The goal is to train a model that performs well for the clients by utilizing both the server data and distributed client data.
We approach the server-client and inter-client domain mismatches by formulating the problem as federated multi-target domain adaptation (FMTDA), illustrated in Figure~\ref{fig:intro}.
Compared to typical FL or domain adaptation (DA), FMTDA introduces several new challenges.
First, it prohibits the use of raw client data on the server.
Most existing DA methods are not applicable to FMTDA since they directly access the target domain data by design.
Second, the inter-client domain discrepancies increase the difficulty to perform FedAvg~\cite{mcmahan2017communication}, a de facto algorithm in FL to aggregate client models on the server.
Since each client model carries the idiosyncrasies of a distinct target domain, averaging the client models tends to cancel out or negatively affect the adaption efforts made on the client devices.
Third, unlike the large-scale and representative target domain data in common DA frameworks, each client owns a limited amount of data.
This implies that the adaptation methods cannot rely on a single client.
Instead, they need to exploit the common properties underlying all client data while acknowledging individual data distribution.
Finally, FMTDA should take into consideration the server-client communication overhead and the imbalanced computational resources.
It is of great interest to develop an algorithm that requires low communication overhead and minimal computation on client devices.
We highlight the aforementioned challenges by ``federating'' several popular DA methods~\cite{ganin2015unsupervised, long2017deep, saito2018maximum}, \ie, modifying them so that they do not access the target domain data on the server end.
Our analyses show that these methods do not account for the inter-client discrepancies, struggle with limited client data, and incur high computational cost on client devices.

To address the issues, we propose a \emph{dual adaptation} (\ours) approach which decouples the training framework into two parts: local adaptation on the client devices and global adaptation on the server.
%
%
On a client device with low computational resources, we freeze the feature extractor in a deep neural network and learn a lightweight local classifier to capture the target characteristics.
Meanwhile, we fit a parametric Gaussian mixture model (GMM) on the client data to encode its statistical distribution.
The GMM parameters and local classifier respectively carry generative and discriminative information of the target domain, which are then uploaded to the server for the computationally heavy feature adaptation.
%
%
On the server side, we jointly update the feature extractor and a global classifier after receiving the adapted local classifiers.
%
Since the server cannot access client data, we design a proxy set to approximate the data distribution of target domains.
We treat the source domain dataset as basis and apply a mixup~\cite{zhang2017mixup} approach to construct a diverse and large-scale domain.
The mixup set provides a wide support that covers the target domains, over which we can weight the instances by a GMM fitted on client data to approximate the corresponding target domain.
%
%
To evaluate the model performance, we conduct extensive experiments on three image classification tasks and two semantic segmentation tasks.
We compare our method with multiple centralized or federated DA baselines in terms of accuracy, communication overhead, and computational cost on client devices.
The main contributions of this work are as follows:
\begin{compactitem}
    \item We introduce a new problem setting, FMTDA, for practical FL vision tasks. 
    It deals with the domain gaps between the unlabeled, distributed client data and a labeled, centralized dataset on the server. 
    \item We identify the key challenges in FMTDA and emphasize them by showing the degraded performance of prior DA methods as well as their ``federated'' version.
    %
    \item We propose the \ours approach to address the new challenges. The main idea is to self-train the local classifiers on client devices and adapt the heavy feature extractor on the server without accessing client data.
    %
    %
    %
%
\end{compactitem}
\vspace{-1mm}
\section{Related Work}
\vspace{-1mm}
%
Peng~\etal~\cite{peng2019federated} introduce a multi-source-single-target DA problem in the FL setting, which is closely related to our work but deals with an opposite scenario.
They assume that the client data is labeled and the unlabeled serve data is available on all local devices.
With this setting, the whole model can be simultaneously trained on local devices but it poses heavy communication and computational demands on the clients.
As such, this method cannot be applied to FMTDA for practical vision tasks. 
To the best of our knowledge, Peng~\etal~\cite{peng2019federated} are the first to study visual domain adaptation in FL, and our work is the first to formalize FMTDA with the consideration of memory, communication, and computational costs.

\vspace{-1mm}
\subsection{Unsupervised domain adaptation}
\vspace{-1mm}
Unsupervised Domain Adaptation (UDA) addresses the domain mismatch problem between labeled source data and unlabeled target data.
Numerous UDA methods have been proposed to transfer the knowledge learned from a source domain to a target domain via centralized training.
Existing approaches can be broadly categorized into divergence-based~\cite{ben2010theory, ghifary2014domain, tzeng2014deep, sun2016deep, long2017deep, rozantsev2018beyond, bhushan2018deepjdot, kang2019contrastive}, reconstruction-based~\cite{ghifary2016deep, yi2017dualgan, zhu2017unpaired, kim2017learning, hoffman2018cycada}, domain adversarial~\cite{ganin2015unsupervised, liu2016coupled, tzeng2017adversarial, liu2018unified}, classifier discrepancy~\cite{saito2018maximum, lee2019sliced}, and self-training~\cite{zhu2017unpaired, zou2019confidence} methods.
In~\cite{ben2010theory}, Ben-David~\etal introduce a divergence criteria to evaluate the domain shift and provide a generalization error bound for domain adaptation.
Ghifary~\etal~\cite{ghifary2016deep} propose deep reconstruction-classification networks to address domain adaptation by learning an additional reconstruction task.
Based on adversarial learning, Ganin and Lempitsky~\cite{ganin2015unsupervised} design a gradient reversal layer to train a domain discriminator, which is widely used in the domain adaptation literature.
Saito~\etal~\cite{saito2018maximum} propose the MCD method to align source and target features by maximizing the discrepancy of two classifiers.
Another group of methods utilize a model trained on source data to generate pseudo label on the target data in a self-training (ST) manner.
Instead of assuming that the target data comes from a single domain, Yu~\etal~\cite{yu2018multi} and Gholami~\etal~\cite{gholami2020unsupervised} address a challenging multi-target UDA problem.
However, these methods are not directly applicable to the FMTDA setting since they assume centralized data on a server.
In addition, reconstruction-based and domain adversarial approaches require additional network modules such as domain discriminators or encoder-decoder architectures, which increase computational cost if applied to the client devices.

\vspace{-1mm}
\subsection{Federated learning}
\vspace{-1mm}
Federated learning~\cite{bonawitz2017practical, mohassel2017secureml, smith2017federated, mohassel2018aby3} is proposed to train a model in a distributed and privacy-preserving manner.
The decentralized learning approaches enable multiple clients to collaboratively learn a model while keeping the training data on local devices.
GiladBachrach~\etal~\cite{gilad2016cryptonets} propose CryptoNets to improve FL performance by enhancing the efficiency of data encryption.
In \cite{bonawitz2017practical}, Bonawitz~\etal introduce a secure aggregation method to update the machine learning models.
Mohassel and Zhang~\cite{mohassel2017secureml} propose SecureML to support privacy-preserving collaborative training in a multi-client FL system.
These methods mainly aim to learn a single global model that works well on centralized data within a single domain.
Recently, several methods are proposed to address the non-{\it i.i.d} distribution of client data.
Smith~\etal~\cite{smith2017federated} introduce federated multi-task learning, which learns a separate model for each node.
In~\cite{liu2018secure}, Liu~\etal propose semi-supervised transfer learning in a privacy-preserving setting.
Hsu~\etal~\cite{hsu2019measuring, hsu2020federated} analyze the effect of non-{\it i.i.d} client data in classification tasks and propose to improve the federated aggregation methods.
On the other hand, Yu~\etal~\cite{yu2020salvaging} focus on the model performance on individual client data and train personalized models via local adaptation.
Nonetheless, the models discussed above require full or semi-supervision on the client data.
In~\cite{jeong2020federated}, Jeong~\etal deal with a disjoint FL scenario where server data is labeled and client data is partially labeled or unlabeled, but the server-client and inter-client domain gaps are not addressed.

\begin{figure*}[t!]
\centering
    \includegraphics[width=.95\linewidth]{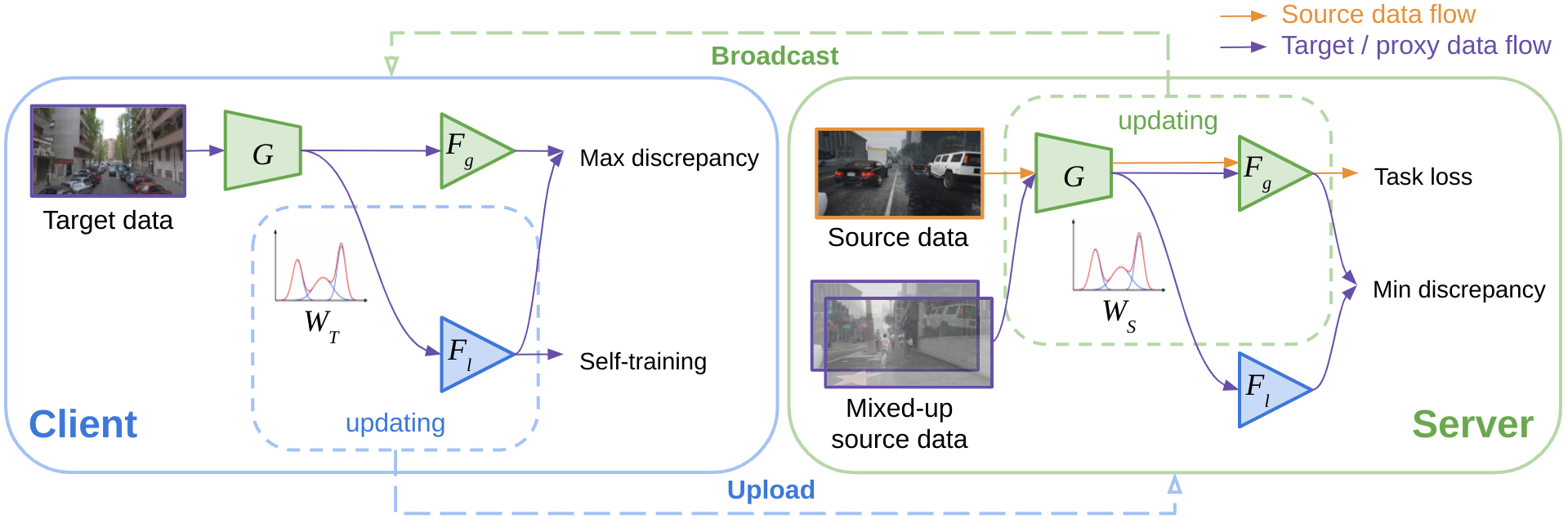}
    \vspace{1mm}
    \caption{\textbf{\ours framework overview} (best viewed in color).
    Our model consists of a feature extractor $G$, a global classifier $F_g$, and a local classifier $F_l$ for each client.
    $G$ and $F_g$ are updated on the server and broadcast to the clients.
    Each client then updates its $F_l$ during local adaptation.
    %
    We also fit a GMM model on both the target $(W_T)$ and source $(W_S)$ data features to weight the examples for training.
    }
\label{fig:framework}
\vspace{-2mm}
\end{figure*}

\vspace{-1mm}
\section{Approach}
\vspace{-1mm}
%
%
Given a centralized, labeled dataset on the server as source domain $D_S$ and the unlabeled data from multiple clients as target domains $({D_T}^i, \ldots, {D_T}^N)$, our goal is to train a model that performs well on the target data.
Unlike centralized DA, the server cannot access the client data due to privacy concerns.
We further prohibit the use of server dataset on client devices considering the limitation of local storage, computation, and communication.
These constraints lead to a multi-party learning framework where all participants keep their data private and only interact via model parameters.
While the server and client models share the same input and output space, their data exhibit domain discrepancies due to various client characteristics.
%

\vspace{-1mm}
\subsection{Preliminaries}
\vspace{-1mm}
In order to minimize the computational cost on client devices, we do not consider resource-consuming DA modules such as adversarial domain discriminators~\cite{ganin2015unsupervised, liu2016coupled, tzeng2017adversarial, liu2018unified} or encoder-decoder architectures~\cite{ghifary2016deep, yi2017dualgan, zhu2017unpaired, kim2017learning, hoffman2018cycada}.
Instead, we exploit maximum classifier discrepancy (MCD)~\cite{saito2018maximum}, which only requires one additional classifier for domain alignment.
A deep neural network classifier can be viewed as a classification head $F$ stacking on top of a feature extractor $G$.
%
%
MCD introduces a second classifier to identify the target data in $D_T$ that is excluded by the support of source domain $D_S$.
The objective of the classifiers $F_1, F_2$ can be expressed as:
\begin{align}
    \min_{F_1,F_2}\quad \mathcal{L}_\text{ce}(D_S)-\mathcal{L}_\text{adv}(D_T) , \label{eq:discrepancy}
\end{align}
where $\mathcal{L}_\text{ce}(D_S)$ is a cross-entropy loss for classifying the labelded source data, and $\mathcal{L}_\text{adv}(D_T)=\mathbb{E}_{\bm{x}\sim D_T}\|F_1(G(\bm{x}))-F_2(G(\bm{x}))\|_1$ is the discrepancy between the  classifier predictions over target data.
Next, the feature extractor $G$ is updated to align the classifier discrepancy as:
\begin{align}
    \min_G\quad \mathcal{L}_\text{adv}(D_T). 
    \label{eq:align}
\end{align}
%
%
These training steps are applied alternately and repeated until convergence.

\vspace{-1mm}
\subsection{Dual adaptation for FMTDA}
\vspace{-1mm}
The decoupling of classifier and feature extractor in MCD makes it a natural fit for FMTDA.
Since it is relatively lightweight to adapt a classifier to the target domain (Eq.~\ref{eq:discrepancy}), we assign it to the client devices and update the feature extractor $G$ on the server (Eq.~\ref{eq:align}).
In addition to the computational efficiency, this design also effectively reduces the communication overhead.
The server broadcasts the whole model to the clients, while the clients only need to upload the classifiers to the server.
%
%
Nonetheless, in this vanilla MCD method, the clients need access to the source data to update $F_1, F_2$ (Eq.~\ref{eq:discrepancy}) and the server requires the target data to adapt $G$ (Eq.~\ref{eq:align}). 
Neither of the requirements satisfies the FMTDA constraints.
%
%
More importantly, each client in FMTDA observes a distinct source-target domain pair due to the non-{\it i.i.d.} data of different clients.
Simply averaging the feature extractors and classifiers on the server can easily lead to canceling effect or negative transfer. 
We address the issue by designating a unique local classifier ${F_l}^i$ to each client $i$.
Meanwhile, we maintain a global classifier $F_g$ to account for the discriminative properties shared by the source and all target domains.
To tackle the data privacy constraints, we propose a self-training approach on the client end and a proxy target domain on the server.
The \ours framework is illustrated in Figure~\ref{fig:framework}.

\vspace{-1mm}
\subsubsection{Self-training local classifiers on the clients.}
\vspace{-1mm}
%
%
In Eq.~\ref{eq:discrepancy}, the term $\mathcal{L}_\text{ce}(D_S)$ is meant to preserve the discriminative power of classifiers and avoid trivial solutions caused by the discrepancy term $\mathcal{L}_\text{adv}(D_T)$.
However, it is infeasible to update the local classifiers $\{{F_l}^i\}$ on client devices since minimizing $\mathcal{L}_\text{ce}(D_S)$ requires the source domain data.
We propose the following variations to self-train classifiers without copying the source data to client devices.
First, we freeze the global classifier $F_g$ and only adapt the local classifier ${F_l}^i$ for client $i$.
As a result, the domain discrepancy loss for client $i$ only applies to the local classifier ${F_l}^i$, which can be written as:
\begin{align}
\mathcal{L}_\text{adv}^i(\bm{x}) = \|F_g(G(\bm{x})) - {F_l}^i(G(\bm{x}))\|_1 ,
\label{eq:mcd}
\end{align}
%
%
%
It not only reduces the communication and local computational costs but stabilizes the local adaptation process.
%
%
%
Second, we preserve the discriminativeness of local classifiers via a self-training loss $\mathcal{L}_\text{st}$.
%
%
It is defined as the cross-entropy between the prediction of the local classifier ${F_l}^i$ and the pseudo label generated by the global classifier ${F_g}$.
%
%
The pseudo label $\hat{y}$ of class $c$ of input $\bm{x}$ is given as:
\begin{equation}
\hat{y}_c (\bm{x}) = 
    \begin{cases}
        1, & \argmax_{k} [{F_{g}} (G(\bm{x}))]_{k} = c \\
        0, & \text{otherwise}
    \end{cases} ,
\label{eqn:st2}
\end{equation}
where $[\bm{p}]_{k}$ is the $k$-th element of a probability vector $\bm{p}$.
%
%

Since the prediction of $F_g$ may not be perfect due to domain mismatch, we use a GMM model fitted on the source data to weight the target examples.
Intuitively, we give more confidence to the prediction of $F_g$ if the example lies closer to the source distribution.
Thus, we send the source GMM parameters $W_S$ from the server to each client.
For each example $\bm{x}$ in the target domain, we calculate the GMM probability $W_S(\bm{x})$ as the confidence of $F_g$ and weight the self-training loss accordingly. 
%
%
%
%
%
%
%
%
%
The local optimization objective for client $i$ can be expressed as:
\begin{equation}
    \min \limits_{{F_l}^i} \hspace{2mm}
    \mathbb{E}_{\bm{x}\sim {D_T}^i} 
    - \mathcal{L}_\text{adv}^i (\bm{x}) + 
    \hspace{1mm} \lambda_\text{st} \hspace{1mm} W_S(\bm{x}) \hspace{1mm} \mathcal{L}_\text{st}^i (\bm{x}) ,
\label{eqn:client}
\end{equation}
where $\lambda_\text{st}$ is a weighting hyper-parameter.
Note that the feature extractor $G$ and global classifier $F_g$ are both fixed on the client devices.
Therefore, the local computation and uploading costs are lightweight as both involve the local classifier ${F_l}^i$ only.
%
%
In addition to the local classifier that captures the discriminative properties of client data, we fit a GMM to describe its generative statistics, thereby allowing the server to construct a proxy set to the client data.
After local adaptation, each client uploads its local classifier $F_l^i$ and GMM parameters ${W_T}^i$ to the server.

\vspace{-1mm}
\subsubsection{Feature alignment via mixup on the server.}
\vspace{-1mm}
%
%
Ideally, the server gathers the local classifier and data from the clients and adapts the feature extractor $G$ as described in Eq.~\ref{eq:align}.
However, the server cannot access target data in the FL settings.
To align features from two domains, we need a certain proxy to approximate the target distribution using only the server data.
In addition to widely-used image transformations, e.g., flipping, cropping, and color jittering, we construct a proxy of the target domains by re-weighing the mixup~\cite{zhang2017mixup} of abundantly available server data.
Mixup is developed to regularize neural network training by densely sampling the convex combinations of training examples.
The convex hull of large-scale source data likely overlaps with the support of target domains considerably.
Moreover, the relationship between empirical risk minimization and mixup is derived~\cite{zhang2017mixup}
and empirically shown to effectively 
improve model generalization, robustness to adversarial examples, and training stability.
These properties are of particular importance in FMTDA since decentralized training with non-{\it i.i.d.} and unlabeled client data tends to be unstable.
By fitting a GMM on each target domain, we can further sample a proxy set according to the data density of each client.
%
%
Note that GMM encodes global statistics in the feature space, which carry less private information than the parameters/gradients of a client model.
%
%
%
%

Specifically, we randomly average two source instances $\bm{x}_m,\bm{x}_n\sim D_S$ in a data batch as a mixup instance $\bm{x}_{mn}=(\bm{x}_m+\bm{x}_n)/2$.
%
%
%
%
Given the GMM parameters ${W_T}^i$ from client $i$, the server uses it to weight each mixup example $\bm{x}_{mn}$, denoted by ${W_T}^i(\bm{x}_{mn})$.
The objective for the server-side adaptation is a weighted average over the mixup examples:
\begin{align}
    \min \limits_{G} \hspace{1mm}
    \sum_{i=1}^N  \mathbb{E}_{\bm{x}\sim \text{mixup}(D_S)} {W_T}^i (\bm{x}) \hspace{1mm} \mathcal{L}_\text{adv}^i(\bm{x}),
\label{eqn:mcd_server}
\end{align}
where $N$ is the total number of clients.
The weighting mechanism allows the feature extractor $G$ to focus on the mixup instances that are closer to a target domain when adapting the extracted features.

%

\vspace{-1mm}
\subsection{Model training and inference}
\vspace{-1mm}
%
Algorithm~\ref{algorithm} describes the algorithmic details of the \ours method and Figure~\ref{fig:framework} illustrates the main  steps. 
We first pre-train the server model ($G$, $F_g$, $W_S$) on the labeled source data $D_S$.
Then, the model is broadcast to all clients and each local classifier ${F_l}^i$ is initialized by $F_g$.
After local optimization (Eq.~\ref{eqn:client}), each client uploads its local classifier and GMM parameters $W_T$ to the server. 
The server first updates the feature extractor and global classifier (Eq.~\ref{eqn:mcd_server}) and then fine-tune the model $(G,F_g)$ using a cross-entropy loss over the source data before broadcasting them back to the clients.
When fitting the GMM models, we reduce the feature dimension with PCA to preserve at least 80\% of the original energy and empirically choose the number of mixture components as twice the number of classes.
The client models are trained with mini-batch gradient descent and the server model is updated by a momentum optimizer to stabilize the training.
%
%
%
%
In the inference phase, we ensemble the predictions of global and local classifiers as:
\begin{equation}
    \tilde{y} (\bm{x}) = 
    ({F_g} (G(\bm{x})) + {F_l}^i (G(\bm{x}))) / 2,
\label{eqn:inference}
\end{equation}
where $\bm{x}$ is a test example from client $i$. 
%
%

\setlength{\textfloatsep}{12pt}
\begin{algorithm}[pt]
    \caption{\hspace{1mm} {\bf \ours}} \label{algorithm}
    \footnotesize
    \begin{flushleft}
        \textbf{Input:} Source domain $D_S = \{(\bm{x_s}, y_s)\}$, 
                         target domains $({D_T}^1, ..., {D_T}^N) = (\{\bm{{x_t}^1}\}, ..., \{\bm{{x_t}^N}\})$,
                         Number of client iterations $R_c$, Number of server iterations $R_s$\\
        \textbf{Output:} Feature extractor $G$, global classifier $F_g$, local classifiers $({F_l}^1, ..., {F_l}^N)$
    \end{flushleft}
    \vspace{-2mm}
    \begin{algorithmic}[1]
        \State Pre-train $G$ and $F_g$ on server with cross-entropy loss
        \Repeat
        \State Fit global GMM ${W_S}$ on source data
        \State Broadcast server models $G, F_g, W_S$ to each client
        \State {\bf \# Local adaptation for client $i$ :}
        \State Initialize local classifier ${F_l}^i \gets F_g$ 
        \State Initialize local GMM ${W_T}^i$
        \For {$r = 1:R_c$} {}
            \State Sample mini-batch $\bm{{x_t}}^i$ from ${D_T}^i$
            \State Update ${F_l}^i$ on $\bm{{x_t}}^i$ with Eq.~\ref{eqn:client}
            \State Update ${W_T}^i$ on $\bm{{x_t}}^i$
        \EndFor
        \State Upload ${F_l}^i$ and ${W_T}^i$ to the server
        \State {\bf \# Server optimization:}
        \For {$r = 1:R_s$} {}
            \State Sample mini-batch $(\bm{x_s}, y_s)$ from $D_S$
            \State Generate mixup data $\bm{x_{s'}}$ from $\bm{x_s}$
            \State Update $G$ on $\bm{x_{s'}}$ with Eq.~\ref{eqn:mcd_server}
            \State Fine-tune $G$ and $F_g$ on $(\bm{x_s}, y_s)$ with cross-entropy loss 
        \EndFor
        \Until {convergence}
    \end{algorithmic}
\label{alg:ours}
\end{algorithm}

\setlength{\textfloatsep}{16pt}
\begin{figure*}[t]
    \centering
    \includegraphics[width=.9\linewidth]{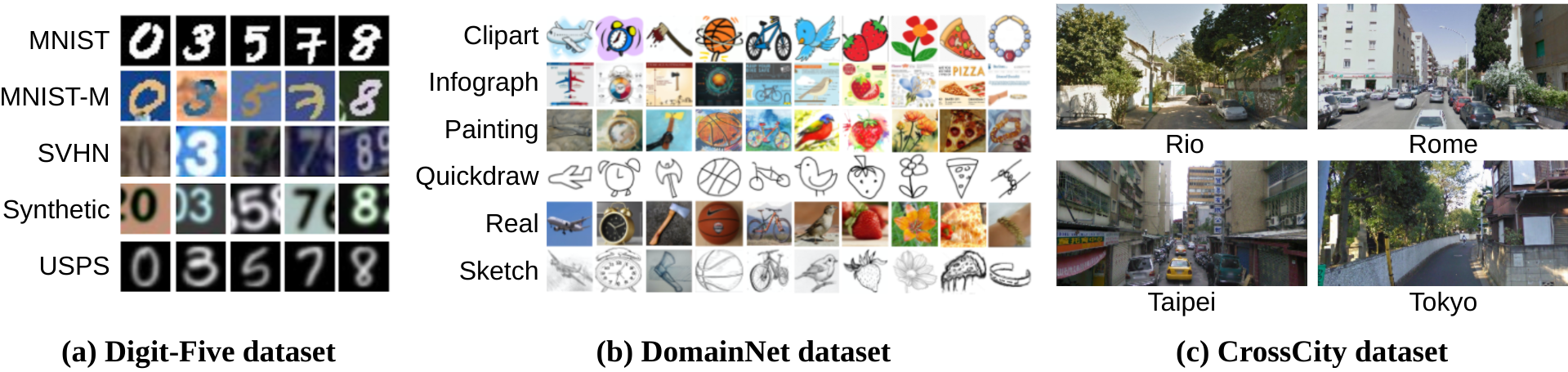}
    \vspace{1mm}
    \caption{\textbf{We evaluate the model performance on three different datasets: Digit-Five, DomainNet, and CrossCity.}
    First, we adapt from MNIST to four other datasets for digit classification.
    %
    %
    Second, we take turn adapting from one image modality in DomainNet to the rest.
    Finally, we perform cross-city adaptation in semantic segmentation from the synthetic GTA5 dataset to real-world street images.
    %
    }
\label{fig:exp}
\end{figure*}

\vspace{-1mm}
\section{Experiments and Analyses}
\vspace{-1mm}

\noindent \textbf{Datasets.}
We experiment with three FMTDA tasks: 
digit classification on 
the Digit-Five dataset~\cite{peng2019federated},
image classification on 
the DomainNet dataset~\cite{peng2019moment}, 
and semantic segmentation with 
adaptation from the synthetic GTA5 dataset~\cite{richter2016playing} to the real-world CrossCity dataset~\cite{chen2017no}.  
Figure~\ref{fig:exp} shows some sample images from these datasets.
Digit-Five is composed of five benchmarks for digit recognition: MNIST~\cite{lecun1998gradient}, Synthetic Digits~\cite{ganin2015unsupervised}, MNIST-M~\cite{ganin2015unsupervised}, SVHN~\cite{netzer2011reading}, and USPS.
The DomainNet dataset contains 596K images of 345 classes from 6 modalities: clipart, inforgraph, painting, quickdraw, real, and sketch.
Each modality is viewed as a domain in this paper.
The GTA5 dataset contains 25k street-view images simulated from computer games.
It provides dense pixel-wise labels for semantic segmentation with 19 classes.
The CrossCity dataset consists of real-world street scenes collected from four different cities: Rio, Rome, Taipei, and Tokyo. 
There are 3.2k training images and 100 testing images for each city.
Since the CrossCity dataset is labeled with only 13 classes, we train and evaluate the models using the overlapped 13 classes between GTA5 and CrossCity.
To simulate a practical scenario, we assume that each client (target domain) only possesses a limited amount of data.
The main results presented here use 10\% data from each target domain in the large-scale Digit-Five and DomainNet datasets, and the supplementary material contains results of other settings.
%
%

\vspace{1pt}
\noindent\textbf{Baselines and upper bounds.}
Since FMTDA is new in the literature, we mainly evaluate our approach, \ours, with the following methods:
1) centralized models trained on the source data only, 
2) centralized MCD~\cite{saito2018maximum} (Cent-MCD) using centralized training in a multi-target DA setting, 
3) federated versions of several DA methods (Fed-DAN~\cite{long2017deep}, Fed-DANN~\cite{ganin2015unsupervised}, Fed-MCD~\cite{saito2018maximum}), and 
4) federated oracle (Fed-oracle), which assumes that \ours can access the target data on the server.

\vspace{1pt}
\noindent\textbf{Implementation details.}
We use the standard FL protocol to extend the centralized DA methods to our setting, i.e., updating models locally and aggregating them on the server using the Federated Averaging (FedAvg) algorithm.
Note that the federated DA baselines require a copy of the source dataset for every client, which gives these schemes some advantage over our method.
The federated oracle model replaces the mixup data on the server with the target domain examples and discards the GMM weighting mechanism.
We implement our model using TensorFlow~\cite{abadi2016tensorflow} and will make the source code available to the public. 
The supplementary material contains more implementation details, model architecture, and additional experimental results on two other adaptation tasks: image classification on the Office-Caltech10 dataset~\cite{gong2012geodesic} and semantic segmentation on the BDD100k dataset~\cite{yu2020bdd100k}.
%

\begin{table}[t]
\centering
\caption{\textbf{Quantitative evaluations on the Digit-Five dataset.}
We report the classification accuracy (\%) on the target domains.
The MCD method produces a clear accuracy gain from the source-only baseline but does not perform well in the federated setting.
Our method improves the performance significantly compared to the federated baselines.
}
\vspace{2mm}
\footnotesize
\setlength\tabcolsep{5pt}
\begin{tabular}{l ccccc}
\toprule
    Method & MNIST-M & SVHN & Synthetic & USPS & Avg
    \\
\midrule
    Source only & 
    26.1 & 10.4 & 26.9 & 57.7 & 30.3
    \\
    Cent-MCD~\cite{saito2018maximum} & 
    28.5 & 12.3 & 29.2 & 70.8 & 35.2
    \\
    Fed-oracle & 
    28.1 & 12.0 & 28.3 & 69.5 & 34.5
    \\
\midrule
    Fed-DAN~\cite{long2017deep} & 
    26.6 & 10.4 & 27.1 & 59.4 & 30.9
    \\
    Fed-DANN~\cite{ganin2015unsupervised} & 
    26.9 & 10.6 & 27.8 & 59.9 & 31.3
    \\
    Fed-MCD~\cite{saito2018maximum} & 
    27.2 & 10.7 & 27.4 & 61.6 & 31.7
    \\
    \ours~(ours) & 
    \bf{27.7} & \bf{11.9} & \bf{28.0} & \bf{68.9} & \bf{34.1}
    \\
\bottomrule
\end{tabular}
\label{tab:digits}
\end{table}

\vspace{-1mm}
\subsection{Task accuracy}
\begin{figure}[t]
    \centering
    \includegraphics[width=.99\linewidth]{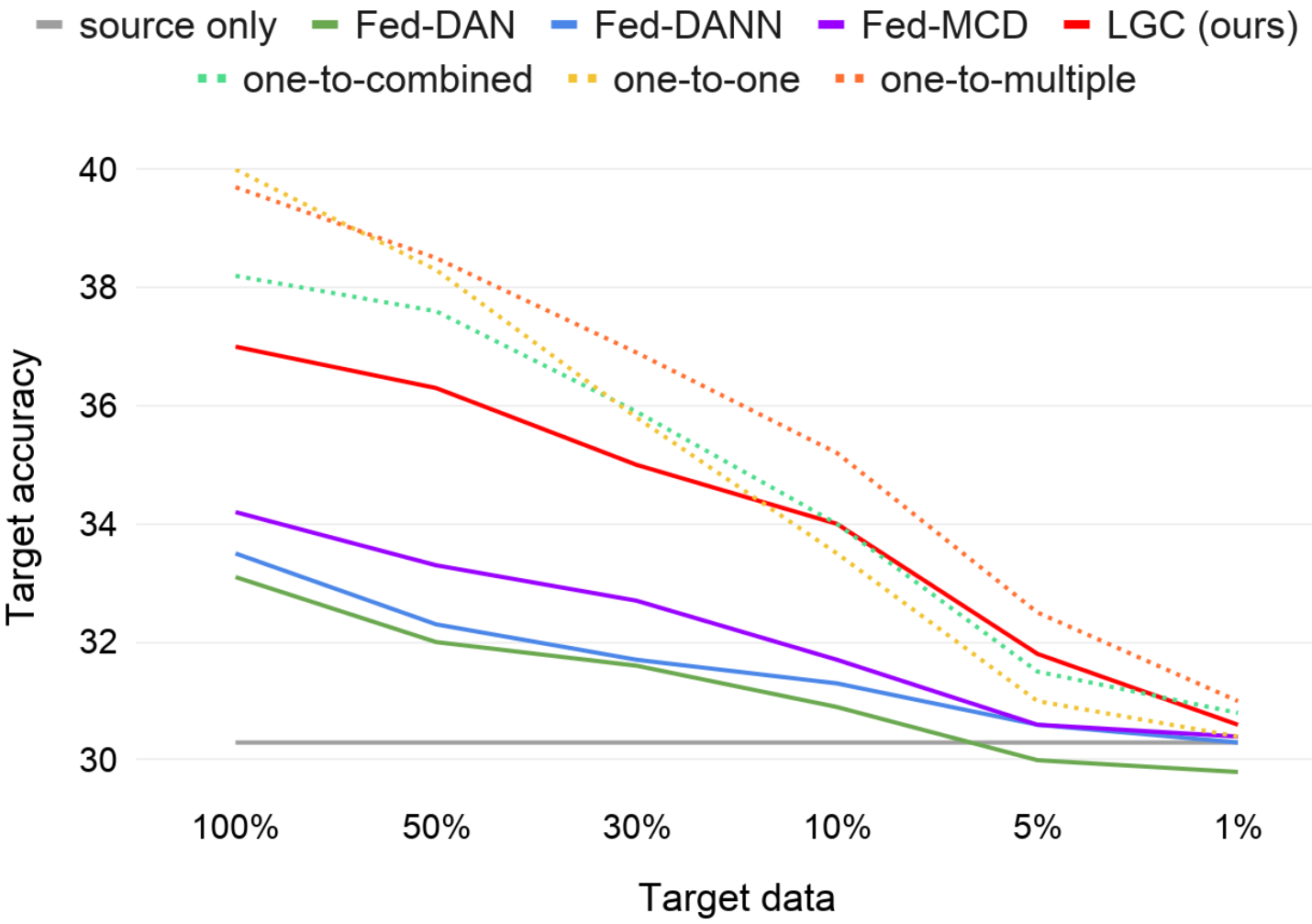}
    \vspace{1mm}
    \caption{\textbf{Quantitative evaluations of different DA methods on the Digit-Five dataset.}
    The solid lines show the federated DA methods and the dash lines are the centralized MCD~\cite{saito2018maximum} method under different training scenarios.
    We evaluate the models trained with various amount of data per client.
    Our model achieves consistently higher accuracy compared to the federated baselines, and can perform better than several centralized methods when the target data on each client device is limited.}
\label{fig:digits}
\end{figure}

\vspace{-1mm}
\noindent\textbf{Digit-Five.}
%
%
%
We use the MNIST dataset as the source domain and the rest as the target domains (i.e., one-to-four adaptation task).
Our models include 4 convolutional layers for feature extraction and two fully-connected layers for each classifier.
%
%
Table~\ref{tab:digits} reports the classification accuracy on the target domains and Figure~\ref{fig:digits} shows the results with various amount of target data for training.
Compared to the source-only baseline, Cent-MCD improves the classification accuracy considerably (from $30.3\%$ to $35.2\%$), confirming the efficacy of domain adaptation. 
Both Cent-MCD and Fed-oracle access the client data on the server, which serve as the upper bounds of our method.
We emphasize that \ours can almost match the performance of Fed-oracle ($34.1\%$ vs.\ $34.5\%$ on average).
%
%
In the FMTDA setting, \ours alleviates the inter-client and source-target domain mismatches.
While the other federated DA methods additionally have access to the source data on client devices, they only tackle the source-target domain discrepancy.
The results show that \ours achieves much higher performance ($34.1\%$) than the federated DA methods ($30.9\%$, $31.3\%$, and $31.7\%$), demonstrating the necessity of modeling the inter-client domain gaps.
%

\begin{table*}[!ht]
\centering
\caption{\textbf{Quantitative evaluations on the DomainNet dataset.}
We take turn using one domain as source and the rest as targets and report the average accuracy (\%) on the target domains.
Our method (\ours) achieves similar accuracy to the centralized and federated upper bounds, which is significantly higher than the federated baselines.
}
\vspace{1mm}
\footnotesize
\setlength\tabcolsep{10pt}
\begin{tabular}{l ccccccc}
\toprule
    Method & 
    Clipart $\rightarrow$ & 
    Infograph $\rightarrow$ & 
    Painting $\rightarrow$ & 
    Quickdraw $\rightarrow$ & 
    Real $\rightarrow$ & 
    Sketch $\rightarrow$ & 
    Avg
    \\
\midrule
    Source only &
    25.7 & 19.3 & 29.6 & 4.4 & 31.5 & 10.5 & 20.2
    \\
    Cent-MCD~\cite{saito2018maximum} &
    29.6 & 23.4 & 34.1 & 7.2 & 36.1 & 14.2 & 24.1
    \\
    Fed-Oracle & 
    29.2 & 22.9 & 33.8 & 6.6 & 35.8 & 13.5 & 23.6
    \\
\midrule
    Fed-DAN~\cite{long2017deep} &
    25.9 & 19.7 & 30.3 & 4.6 & 31.1 & 11.4 & 20.5
    \\
    Fed-DANN~\cite{ganin2015unsupervised} &
    25.5 & 20.1 & 30.4 & 4.7 & 31.3 & 10.8 & 20.5
    \\
    Fed-MCD~\cite{saito2018maximum} &
    27.1 & 19.9 & 31.4 & 4.7 & 32.5 & 10.4 & 21.0
    \\
    \ours~(ours) &
    \bf{29.0} & \bf{22.3} & \bf{33.5} & \bf{6.0} & \bf{35.7} & \bf{13.2} & \bf{23.3}
    \\
\bottomrule
\end{tabular}
\label{tab:domainnet}
\end{table*}

\vspace{1pt}
\noindent\textbf{DomainNet.}
We perform cross-modality adaptation on the DomainNet dataset~\cite{peng2019moment} using ResNet101~\cite{He2015} for feature extraction. 
%
Each modality in the DomainNet is taken as the source domain in turn.
The quantitative results are shown in Table~\ref{tab:domainnet}.
%
The performance of evaluated methods are similar to that in the Digit-Five dataset. 
Adapting from ``quickdraw'' and ``sketch'' to the rest are more challenging due to their larger modality gaps than other domains. 
However, \ours is able to achieve consistent performance gains over the other federated DA methods.

\begin{table}[t]
\centering
\caption{\textbf{Quantitative evaluations on GTA5-to-CrossCity.}
In this challenging segmentation task, the federated baselines produce minimal mIoU gain on the target domains whereas our method improves the performance by a clear margin.
}
\vspace{1mm}
\footnotesize
\setlength\tabcolsep{7pt}
\begin{tabular}{l ccccc}
\toprule
    Method & Rio & Rome & Taipei & Tokyo & Avg
    \\
\midrule
    Source only & 
    27.9 & 27.6 & 26.0 & 28.2 & 27.4
    \\
    Cent-MCD~\cite{saito2018maximum} & 
    31.1 & 30.6 & 28.8 & 31.6 & 30.5
    \\
    Fed-oracle & 
    30.3 & 28.4 & 28.3 & 30.7 & 29.4
    \\
\midrule
    Fed-DAN~\cite{long2017deep} & 
    27.3 & 26.4 & 26.0 & 28.5 & 27.1
    \\
    Fed-DANN~\cite{ganin2015unsupervised} & 
    28.6 & 26.0 & 26.6 & 28.6 & 27.5
    \\
    Fed-MCD~\cite{saito2018maximum} & 
    27.7 & 27.3 & 26.5 & 29.0 & 27.6
    \\
    \ours (ours) & 
    \bf{29.2} & \bf{28.0} & \bf{27.6} & \bf{30.7} & \bf{28.9}
    \\
\bottomrule
\end{tabular}
\label{tab:crosscity}
\end{table}

\vspace{1pt}
\noindent\textbf{GTA5 to CrossCity.}
For this semantic segmentation task, the domain adaptation is from the synthetic GTA5 dataset~\cite{richter2016playing} to the four real-world cities in CrossCity~\cite{chen2017no}.
We use the MobileNetv2~\cite{sandler2018mobilenetv2} with multiplier $\alpha=0.5$ as backbone and DeepLabv3~\cite{chen2017rethinking} without decoder as our segmentation model.
%
%
%
%
Although GTA5 is a large-scale dataset, the diverse scene structures across different target cities pose a great challenge for adaptation.
Table~\ref{tab:crosscity} shows the quantitative results of the evaluated methods based on the mean intersection-over-union (mIoU) metric.
While Cent-MCD achieves clear performance gain over the source-only baseline in the centralized setting, the federated adaptation methods do not preform as well in the FL setting.
Our method improves from the baseline by $1.3$ mIoU, approaching the oracle upper bound (28.9 vs.\ 29.4).
%

\begin{table*}[!ht]
\centering
\caption{\textbf{Evaluations of communication and computational costs.}
We calculate the client-end FLOPS as computational cost during training and number of model parameters to upload + broadcast as communication overhead.
Our method (\ours) reduces the communication and computational costs significantly compared to the other baselines.
}
\vspace{1mm}
\footnotesize
\begin{tabular}{l rr rr rr}
\toprule
    & \multicolumn{2}{c}{\bf{Digit-Five}} & 
    \multicolumn{2}{c}{\bf{DomainNet}} & 
    \multicolumn{2}{c}{\bf{GTA5-to-CrossCity}}
    \\
    Method & Computation & Communication &
             Computation & Communication &
             Computation & Communication
    \\
\midrule
    Fed-DAN~\cite{long2017deep} & 
    314.0M & 493K + 493K &
    36.1B & 2.5M + 2.5M &
    31.5B & 45.4M + 45.4M
    \\
    Fed-DANN~\cite{ganin2015unsupervised} & 
    314.3M & 493K + 493K &
    36.1B & 2.5M + 2.5M &
    32.0B & 45.4M + 45.4M
    \\
    Fed-MCD~\cite{saito2018maximum} & 
    314.6M & 510K + 510K &
    36.1B & 2.5M + 2.5M &
    32.6B & 46.1M + 46.1M
    \\
    \ours (ours) & 
    \bf{78.7M} & \bf{18K + 494K} &
    \bf{9.0B} & \bf{0.02M + 2.5M} &
    \bf{8.4B} & \bf{1.6M + 46.2M}
    \\
\bottomrule
\end{tabular}
\label{tab:cost}
\vspace{-1mm}
\end{table*}

\begin{table*}[!ht]
\centering
\caption{\textbf{Ablative evaluations on the Digit-Five dataset.}
We show that \ours significantly reduces the communication and computational costs by decoupling client and server model training.
The individual components of our framework like ST and GMM improve the accuracy and barely require additional communication and computational costs.}
%
%
%
\vspace{1mm}
\footnotesize
\begin{tabular}{lll crr}
\toprule
    Method & Client & Server & Accuracy (\%) & Computation (FLOPS) & Communication (\# parameters)
    \\
\midrule
    Fed-MCD~\cite{saito2018maximum} & MCD target & - & 
    31.7 & 314.6M & 510K + 510K
    \\
    \ours & MCD target & MCD mixup & 
    32.8 & \bf{78.5M} & \bf{17K + 493K}
    \\
    \ours & MCD target + ST & MCD mixup & 
    33.5 & \bf{78.5M} & \bf{17K + 493K}
    \\
    \ours & MCD target + ST & MCD mixup + GMM & 
    34.1 & 78.7M & 18K + 494K
    \\
\midrule
    Fed-oracle & MCD target + ST & MCD target & 
    34.5 & 78.5M & 17K + 493K
    \\
\bottomrule
\end{tabular}
\label{tab:ablation}
\vspace{-1mm}
\end{table*}

\vspace{-1mm}
\subsection{Communication and computational costs}
\vspace{-1mm}
In addition to measuring the task accuracy on target data, we also evaluate the server-client communication and on-device computational costs.
For communication, we calculate the number of model parameters that need to be transmitted in two directions: upload (client to server) and broadcast (server to client).
Regarding the computational cost, we emphasize the importance of low training cost on the client devices since they are usually equipped with limited computational resources.
Specifically, we estimate the number of floating-point operations (FLOPs) incurred by a data example in a forward pass and backpropagation during training.
Without loss of generality, we divide most methods into a feature extractor $G$, a classifier $F$, and optionally a domain discriminator $D$.
For each module $m\in\{G, F, D\}$, we denote by $|m|$ the number of its parameters and by $\Vert m \Vert$ the FLOPs of a forward pass of one data example.
The FLOPs in backpropagation are typically similar to the forward pass FLOPs $\|m\|$ in our experiments, so we estimate the on-device training cost by $2\|m\|$ if the module is trainable.

We use the federated DA baselines and our method to demonstrate how to calculate the communication and on-device computational costs.
Recall that in the federated DA baselines, each client has access to its own target domain data as well as the source data from server.
%
%
%
%
\begin{itemize}[noitemsep,topsep=0pt,parsep=0pt,partopsep=0pt]
\item The DANN method~\cite{ganin2015unsupervised} aligns the source and target domains by inserting a gradient reversal layer between the feature extractor and domain discriminator.
We train a domain discriminator for each client and update it locally with the main network.
During local optimization, each target domain example is passed through $G$ and $D$, and source domain example is passed through all the modules ($G$, $F$, and $D$).
Hence, the on-device computational cost is $2(\Vert G \Vert+\Vert D \Vert)+2(\Vert G \Vert+\Vert F \Vert+\Vert D \Vert)$ in Fed-DANN.
For data communication, each client uploads its $G$ and $F$ to the server, and the server broadcasts the aggregated $G$ and $F$ back to the clients.
The communication overhead is $|G|+|F|$ in both directions.
%
\item Fed-MCD~\cite{saito2018maximum} trains a feature extractor and two classifiers for each federated client.
The computational cost is thus $2(\Vert G \Vert+2\Vert F \Vert)+2(\Vert G \Vert+2\Vert F \Vert)$, and the communication overhead is $|G|+2|F|$.
%
%
%
\item In \ours, the feature extractor and global classifier are fixed on client devices, and we only pass the target data through the local model.
It reduces the computational cost to $\Vert G \Vert+3\Vert F \Vert$, including one forward pass over $G$, $F_g$, and $F_l$ and a backpropagation pass over $F_l$.
For data communication, we upload $|F|+|W|$ parameters and broadcast $|G|+|F|+|W|$ parameters.
\end{itemize}
%
%
%
%

%
%

Generally, the communication and computational costs are dominated by the feature extractor $G$ of a model.
\ours reduces the computational and upload costs considerably compared to the federated baselines as we do not update $G$ on-device and do not pass the source data through the local model.
Fitting GMM models to the target data barely creates additional costs per data example per iteration since the feature dimension is reduced.
In Table~\ref{tab:cost}, we show the communication and computational costs incurred by our method and other federated baselines.
%
%
%
%
%
In all three experiments, \ours requires approximately 1/4 FLOPS of other methods during the client-end training.
Moreover, the number of model parameters that need to be uploaded to the server is reduced considerably.
The results demonstrate that \ours can perform the adaptation efficiently with the minimal costs of communication and computation.

\vspace{-1mm}
\subsection{Ablation studies}
\vspace{-1mm}
To evaluate the effectiveness and costs of individual components in our framework, we perform ablation studies on the Digit-Five experiment.
Table~\ref{tab:ablation} shows, by introducing local and global classifiers, we achieve a $1.1\%$ accuracy gain and require only a quarter of the computational cost and half of the communication overhead compared to Fed-MCD.
Self-training (ST) further improves the accuracy by $0.7\%$ and barely increases the computation overhead.
With the GMM weighting mechanism, the accuracy of our full model is close to the oracle while having only 1K additional parameters to upload and broadcast.
It demonstrates the effectiveness of weighting mixup data as target proxy.

\vspace{-1mm}
\section{Centralized multi-target domain adaptation}
\vspace{-1mm}
%
%
%
%
%
%
Our work is closely related to centralized multi-target DA.
In a typical FL system, the client models are trained on locally collected data which is non-{\it i.i.d.} due to different user characteristics.
The problem of domain mismatch among multiple clients, or target domains, is challenging even in a centralized training framework.
Although some methods have been proposed for centralized multi-target DA~\cite{yu2018multi, gholami2020unsupervised}, the best way to adapt to diverse domains in real-world scenarios remains unclear.
To disentangle the challenges of multi-target DA and distributed training, we also look into the problem in a centralized setting.
The idea is to investigate how the performance of a DA method changes after eliminating the federated setting.

When dealing with multiple target domains, some common solutions include 1) treating the adaptation to each target as a separate DA task (one-to-one), 2) combining all target data into one domain and solving it as single-source-single-target DA (one-to-combined), and 3) adapting to multiple targets simultaneously by decomposing model parameters or feature representation into shared and private components (one-to-multiple).
We evaluate MCD~\cite{saito2018maximum} with these settings using the Digit-Five experiment.
In the one-to-multiple setting, the model is decomposed into a shared feature extractor and two classifiers for each target domain. 
As shown in Figure~\ref{fig:digits} (dash lines), the one-to-one adaptation models perform better when sufficient training data from the target domain is available.
In a practical scenario with fewer target examples, the one-to-multiple method can more effectively adapt to the target domains.
Note that the one-to-one adaptation requires multiple copies of the model parameters while one-to-combined needs only one.
However, the one-to-combined method is not applicable in federated learning since the client data is private.
The observations justify our one-to-multiple FL model design since the federated clients usually possess limited data.

\vspace{-1mm}
\section{Conclusion}
\vspace{-1mm}
Multi-target domain adaptation is a natural and challenging problem in federated learning.
While most existing methods assume labeled and/or {\it i.i.d} client data, we consider a practical setting: multi-target unsupervised domain adaptation.
We show that naively applying centralized DA methods on federated client devices not only leads to poor performance but also costs considerable communication and computational overheads.
To address this, we propose a simple yet effective framework, \ours, which requires minimal FL training costs.
Extensive experiments on image classification and semantic segmentation demonstrate that \ours can achieve significant performance gain compared to centralized and federated baselines.
We hope that this work will encourage more research attention to this novel and crucial topic.

{\small
\bibliographystyle{ieee_fullname}
\bibliography{egbib}
}

\end{document}